\def\year{2021}\relax
\documentclass[letterpaper]{article} 
\usepackage{aaai21}  
\usepackage{times}  
\usepackage{helvet} 
\usepackage{courier}  
\usepackage[hyphens]{url}  
\usepackage{graphicx} 
\usepackage{graphicx}  
\usepackage{natbib}  
\usepackage{caption} 
\usepackage{graphicx}
\usepackage[nottoc]{tocbibind}
\usepackage{amsfonts}
\usepackage{amsmath}
\usepackage{mathtools}
\usepackage{booktabs}
\usepackage{hyperref}

\DeclarePairedDelimiter{\norm}{\lVert}{\rVert}

\title{Self-Supervised Sketch-to-Image Synthesis}

\author{Bingchen Liu$^{1,2}$, Yizhe Zhu$^2$, Kunpeng Song$^{1,2}$, Ahmed Elgammal$^{1,2}$\\
$^1$Playform - Artrendex Inc., USA \\
$^2$Department of Computer Science, Rutgers University\\
{\tt \{bingchen.liu, yizhe.zhu, kungpeng.song\}@rutgers.edu, elgammal@artrendex.com}
}

\begin{document}

\maketitle

\begin{figure*}[h]
    \centering
    \includegraphics[width=0.8\linewidth,height=5.4cm]{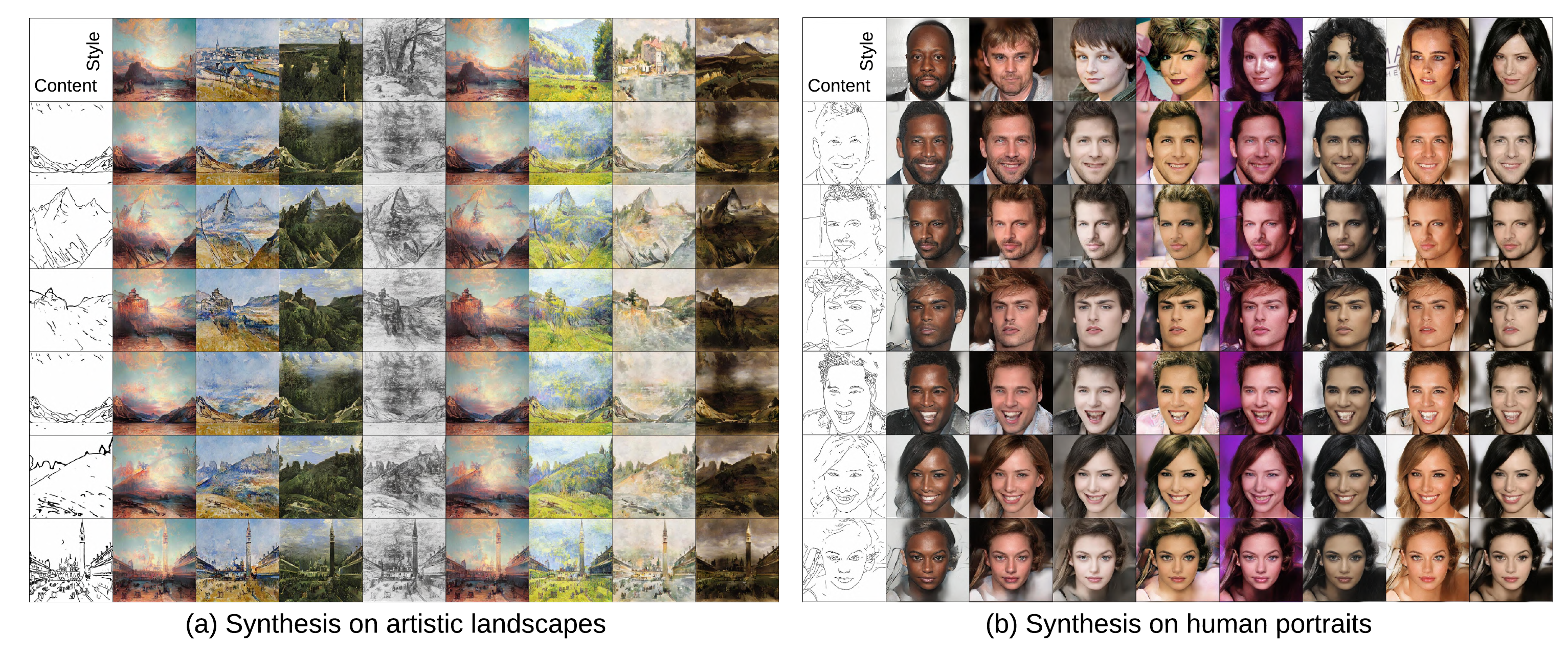}
    \caption{Exemplar-based sketch-to-image synthesis from our model on varied image domains in $1024^2$ resolution.}
    \label{fig:samples-overview}
\end{figure*}

\begin{abstract}
Imagining a colored realistic image from an arbitrary drawn sketch is one of human capabilities that we eager machines to mimic.  Unlike previous methods that either require the sketch-image pairs or utilize low-quantity detected edges as sketches, we study the exemplar-based sketch-to-image (s2i) synthesis task in a self-supervised learning manner, eliminating the necessity of the paired sketch data. To this end,  we first propose an unsupervised method to efficiently synthesize line-sketches for general RGB-only datasets. With the synthetic paired-data, we then present a self-supervised Auto-Encoder (AE) to decouple the content/style features from sketches and RGB-images, and synthesize images that are both content-faithful to the sketches and style-consistent to the RGB-images. While prior works employ either the cycle-consistence loss or dedicated attentional modules to enforce the content/style fidelity, we show AE's superior performance with pure self-supervisions.  To further improve the synthesis quality in high resolution, we also leverage an adversarial network to refine the details of the synthetic images. Extensive experiments on $1024^2$ resolution demonstrate a new state-of-art-art performance of the proposed model on CelebA-HQ and Wiki-Art datasets. Moreover, with the proposed sketch generator, the model shows a promising performance on style mixing and style transfer, which require synthesized images to be both style-consistent and semantically meaningful. Our code is available on \href{https://github.com/odegeasslbc/Self-Supervised-Sketch-to-Image-Synthesis-PyTorch}{GitHub}, and please visit \href{https://create.playform.io/my-projects?mode=sketch}{Playform.io} for an online demo of our model.
\end{abstract}

\section{Introduction}
Exemplar-based sketch-to-image (s2i) synthesis has received active studies recently \cite{liu2019unpaired,zhang2020cross,Lee_2020_CVPR,liu2020sketch} for its great potential in assisting human creative works \cite{elgammal2017can,kim2018finding,elgammal2018shape}. Given a referential image that defines the style, an s2i model synthesizes an image from an input sketch with consistent coloring and textures to the reference style image. A high-quality s2i model can help reduce repetitive works in animation, filming, and video game story-boarding. It can also help in sketch-based image recognition and retrieval. Moreover, since the model generates images that are style-consistent to the referential images, it has great potential in style-transfer and style harmonization, therefore impacting the human artistic creation processes.

Sketch-to-image synthesis is one important task under the image-to-image (i2i) translation \cite{isola2017image,liu2017unsupervised,zhu2017unpaired,kim2019u} category, which benefits a lot from recent year's advances in generative models \cite{kingma2013auto,goodfellow2014generative}. Unlike general i2i tasks, exemplar-based s2i is challenging in several aspects:
1) The sketch domain contains limited information to synthesize images with rich content; especially, real-world sketches have lines that are randomly deformed and differ a lot from the edges in the desired RGB-images.
2) The referential style image usually has a big content difference to the sketch, to avoid content-interference from the style image, the model has to disentangle the content and style information from both inputs effectively.
3) Datasets with paired sketches and RGB-images are rare, even for unpaired sketches that are in the same content domain as the RGB dataset are hard to collect. 

Existing works mostly derive their customized attention modules \cite{vaswani2017attention,zhang2019self}, which learn to map the style cues from the referential image to the spatial locations in the sketch, to tackle the first two challenges, and leverage a cycle-consistent \cite{zhu2017unpaired} or back-tracing \cite{liu2017unsupervised} framework to enforce the style and content faithfulness to the respective inputs. However, the derived attention modules and the required supporting models for consistency-checking significantly increase the training cost and limit them to work on low resolution ($256^2$) images. Moreover, due to the lack of training data, previous methods either work around edge-maps rather than free-hand sketches or on datasets with limited samples, restricting their practicality on image domains with more complicated style and content variance.  

Aiming to break the bottleneck on datasets with reliable matched sketches and RGB-images, we propose a dedicated image domain-transfer \cite{gatys2016image,huang2017arbitrary} model. The model synthesizes multiple paired free-hand sketches for each image in large RGB datasets. Benefit from the paired data, we then show that a simple Auto-encoder (AE) \cite{kramer1991nonlinear,vincent2010stacked} equipped with self-supervision \cite{feng2019self,kolesnikov2019revisiting,he2020momentum} 
exhibits exceptional performance in disentangling the content and style information and synthesizing faithful images. As a result, we abandon commonly-used strategies such as cycle-consistent loss and attention mechanisms. It makes our model neat with less computation cost while having a superior performance at $1024^2$ resolution. 

In summary, our contributions in this work are:
\begin{itemize}
    \item We propose a line-sketch generator for generic RGB-datasets, which produces multiple sketches for one image.
    \item We introduce an efficient self-supervised auto-encoder for the exemplar-based s2i task, with a momentum-based mutual information minimization loss to better decouple the content and style information.
    \item We present two technique designs in improving DMI \cite{liu2020sketch} and AdaIN \cite{huang2017arbitrary}, for a better synthesis performance.
    \item We show that our method is capable of handling both the high-resolution s2i task and the style-transfer task with a promising semantics-infer ability.
\end{itemize}

\section{Related Work}
\textbf{Basics} Auto-encoder \cite{kramer1991nonlinear,vincent2010stacked} (AE) is a classic model that has been widely applied in image-related tasks. Once trained, the decoder in AE becomes a generative model which can synthesize images from a lower-dimensional feature space. Apart from AE, Generative Adversarial Network (GAN) \cite{goodfellow2014generative} significantly boosts the performance in image synthesis tasks. GAN involves a competition between a generator $G$ and a discriminator $D$, where $G$ and $D$ iteratively improves each other via adversarial training. 

\noindent\textbf{Sketch to image synthesis} Recent s2i methods can be divided into two categories by the training scheme they based on 1) Pix2pix-based methods \cite{isola2017image} which is a conditional-GAN \cite{mirza2014conditional} while $G$ is in the form of an encoder-decoder, and paired data is required to train $G$ as an $AE$; 2) CycleGAN-based methods \cite{zhu2017unpaired} that accept unpaired data but require two GANs to learn the transformations back and forth. 

Representing Pix2pix-based models includes AutoPainter \cite{liu2017auto}, ScribblerGAN \cite{sangkloy2017scribbler}, and SketchyGAN \cite{chen2018sketchygan}. However, none of them have a delicate control to synthesis via exemplar-images. Sketch2art \cite{liu2020sketch} addresses style-consistency to a referential image, but requires an extra encoder for style feature extraction. \citeauthor{zhang2020cross} and \citeauthor{Lee_2020_CVPR} propose reference-based module (RBNet) and cross-domain correspondence module (CoCosNet) respectively, both leverage an attention map to relocate the style cues to the sketch, to enable the exemplar-based synthesis.

Early successors of CycleGAN includes UNIT \cite{liu2017unsupervised}, which employs an extra pair of encoders to model an assumed domain-invariant feature space. MUNIT \cite{huang2018multimodal,lee2018diverse} further achieves multi-modal image translation. U-GAT-IT \cite{kim2019u} is a recent exemplar-based model which includes an attention module to align the visual features from the content and style inputs. Furthermore, US2P \cite{liu2019unpaired} is the latest work that dedicates to s2i, which first translates between sketch and grey-scale images via a CycleGAN, then leverages a separate model for exemplar-based coloration.

Different from both categories, only an simple auto-encoder is applied in our model. We show that an AE, with self-supervision methods including data-augmenting and self-contrastive learning, is sufficient to get remarkable content inference and style translation. 

\label{sec:related-work}

\section{Sketch Synthesis for Any Image Dataset}
\label{sec:synthesis-skt}
Few of the publicly available RGB-image datasets have paired sketches, and generating realistic line-sketches for them is challenging. Edge-detection methods \cite{canny1986computational,xie2015holistically} can be leveraged to mimic the ``paired sketches"; however, such methods lack authenticity. Moreover, the lack of generalization ability on edge detection methods can lead to missing or distracting lines. There are dedicated deep learning models on synthesizing sketches \cite{chen2018semi,li2019im2pencil,yu2020toward}, but most of them focus on pencil sketches with domain-specific tweaks (e.g., only works for faces). Instead, we are interested in sketches of simple lines \cite{simo2018mastering} that one can quickly draw, and should be realistic with random shape deformations (lines that are neither straight nor continuous). 

\begin{figure}[h]
    \centering
    \includegraphics[width=0.8\linewidth,height=6.4cm]{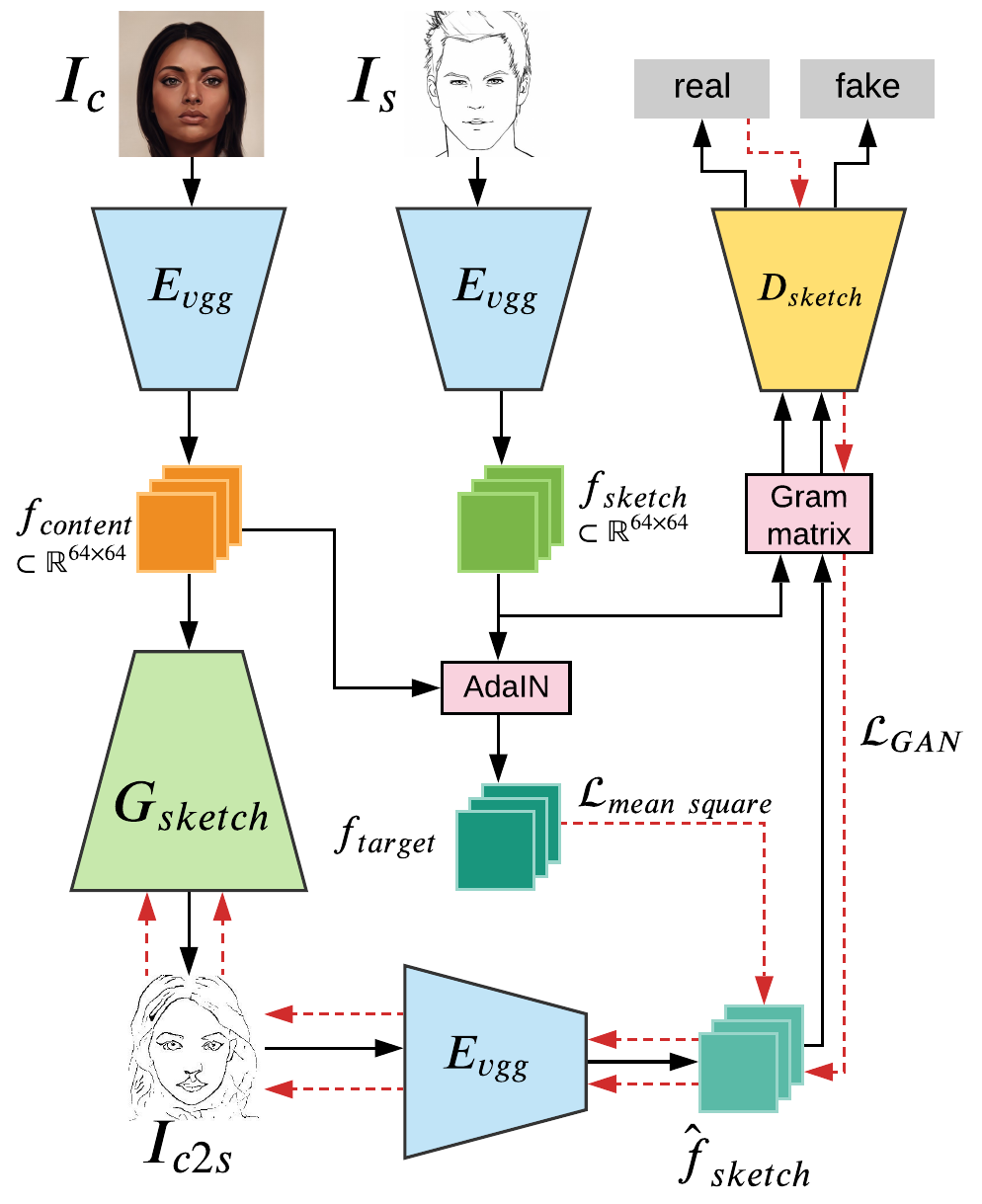}
    \caption{Illustration of our TOM. Dashed arrows in red indicate the gradient flow to train the sketch generator.}
    \label{fig:skt-gen-1}
\end{figure}

We consider the sketch synthesis as an image domain transfer problem, where the RGB-image domain $R$ is mapped to the line-sketch domain $S$.  Accordingly, we propose a GAN-based domain transfer model called TOM, short for ``Train Once and get Multiple transfers". To produce multiple paired sketches for each image in $R$, we design an online feature-matching scheme, and to make TOM neat and efficient, we adopt a single-direction model which we empirically found performing well enough for our sketch generation purpose. We will show that the model is 1) fast and effective to train on $R$ with varied domains, such as faces, art paintings, and fashion apparels, 2) so data-efficient that only a few line-sketches (not even need to be in an associated domain to $R$) are sufficient to serve as $S$.

TOM consists of three modules: a pre-trained VGG \cite{simonyan2014very} $E$ that is fixed, a sketch Generator $G_{sketch}$, and a Discriminator $D_{sketch}$. We have:
\begin{align}
    &f_{content} = E(I_c), \qquad\qquad\qquad I_c \sim \mathcal{R}; \\
    &f_{sketch} = E(I_s),  \qquad\qquad\qquad\; I_s \sim \mathcal{S}; \\
    &\hat{f}_{sketch} = E(I_{c2s}), \qquad\qquad\quad I_{c2s} = G( f_{content} ); \\
    &f_{target} =  \sigma(f_{sketch}) \cdot \mathrm{IN}(f_{content}) + \mu(f_{sketch}),
\end{align}
where $\mathrm{IN}$ is instance normalization \cite{ulyanov2016instance}, and $f_{target}$ is produced via adaptive $\mathrm{IN}$ (AdaIN \cite{huang2017arbitrary}) which possesses the content information of $I_c$ while having the feature statistics of $I_s$. As shown in Figure~\ref{fig:skt-gen-1}, $D_{sketch}$ is trained to distinguish the feature statistics from real sketches $I_s$ and generated sketches $I_{c2s}$. $G_{sketch}$ is trained to synthesis sketches $I_{c2s} = G(E(I_c))$ for an RGB image $I_c$. The objectives of TOM are:
\begin{align}
    \mathcal{L}_{D_{sketch}} = -\mathbb{E}[ log(D_{sketch}( Gram(f_{sketch}) )) ]& \nonumber \\
                     -\mathbb{E}[ log(1 - D_{sketch}( Gram( \hat{f}_{sketch} ) )) ]&, \\
    \mathcal{L}_{G_{sketch}}= -\mathbb{E}[ log( D_{sketch}( Gram( \hat{f}_{sketch} ) )) ]&  \nonumber\\
                     + \mathbb{E}[ \norm{ f_{target} - \hat{f}_{sketch} }^2 ]&,\label{Eq:tom-mse}
\end{align}
where $Gram$ is gram matrix \cite{gatys2016image} which computes the spatial-wise covariance for a feature-map. The objectives for $G_{sketch}$ are two-fold. Firstly, the discriminative loss in Eq.\ref{Eq:tom-mse} makes sure that $I_{c2s}$ is realistic with random deformations and stroke styles, and enables $G_{sketch}$ to generalize well on all images from $\mathcal{R}$. Secondly, the mean-square loss in Eq.\ref{Eq:tom-mse} ensures the content consistency of $I_{c2s}$ to $I_c$.

Importantly, we randomly match a batch of RGB-images $I_c$ and real sketches $I_s$ during training. Therefore, $f_{target}$ is created in an online fashion and is always changing for the same $I_c$. In other words, for the same $I_c$, Eq.\ref{Eq:tom-mse} trains $G_{sketch}$ to generate a sketch towards a new ``sketch style" in every new training iteration. Combined with such an online feature-matching training strategy, we leverage the randomness from the SGD optimizer \cite{robbins1951stochastic} to sample the weights of $G_{sketch}$ as checkpoints after it is observed to output good quality $I_{c2s}$. As a result, we can generate multiple sketches for one image according to the multiple checkpoints, which can substantially improve our primary sketch-to-image model's robustness.

\begin{figure*}[h]
    \centering
    \includegraphics[width=0.86\linewidth,height=5.3cm]{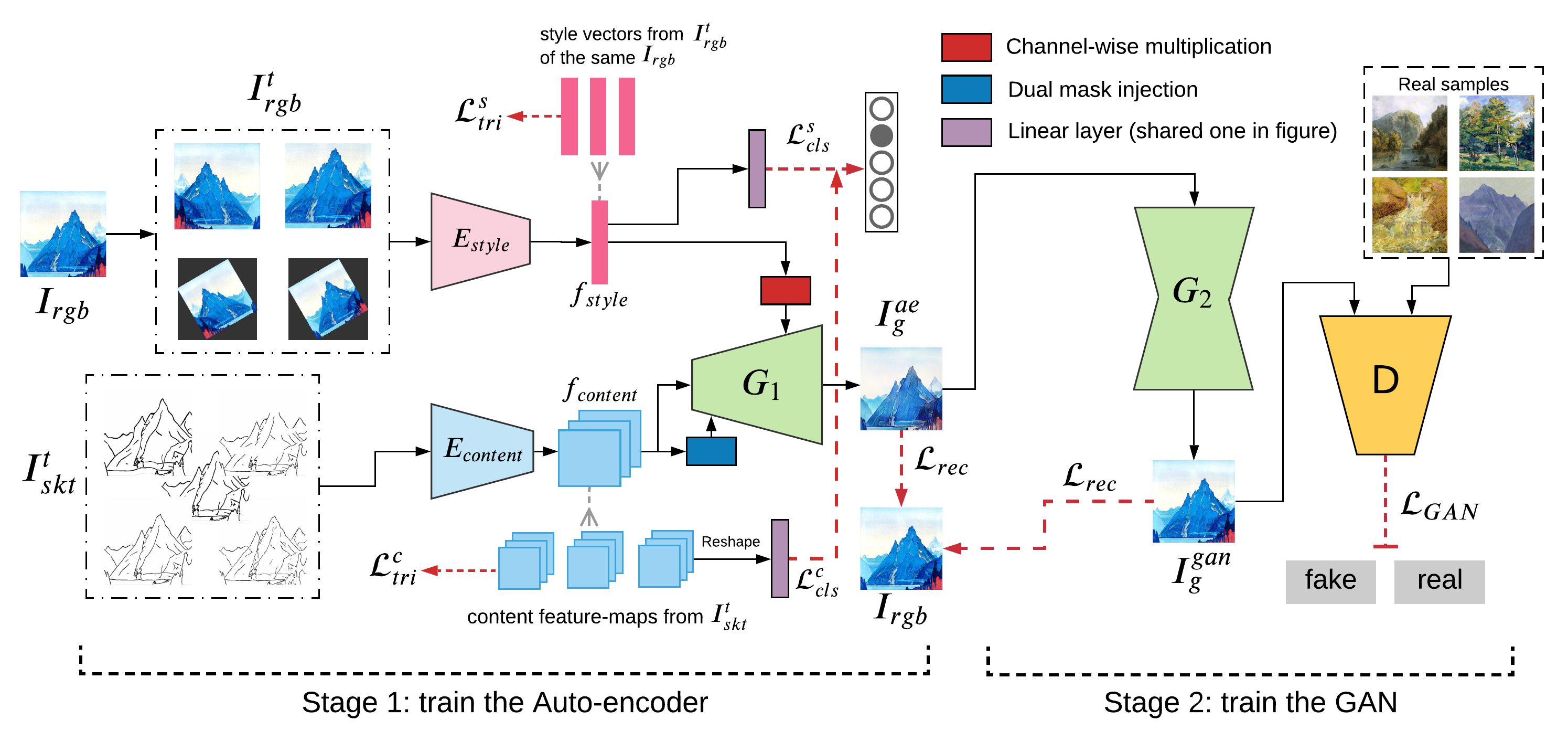}
    \caption{Overview of the proposed model.}
    \label{fig:model-overview}
\end{figure*}

\section{Style-guided Sketch to Image Synthesis}

We consider two main challenges in the style-guided sketch to image synthesis: 1) the style and content disentanglement, 2) the quality of the final synthesized image. We show that with our designed self-supervised signals, an Auto-Encoder (AE) can hallucinate rich content from a sparse line-sketch while assigning semantically appropriate styles from a referential image. After the AE training, we employ a GAN to revise the outputs from AE for a higher synthesis quality.

 \subsection{Self-supervised Auto-encoder}

Our AE consists of two separate encoders: 1) a style encoder $E_{style}$ that takes in an RGB-image $I^{t}_{rgb}$ to generate a style vector $f_{style} \subseteq \mathbb{R}^{512}$, 2) a content encoder $E_{content}$ which takes in a sketch $I^{t}_{skt}$ and extracts a content feature-map $f_{content} \subseteq \mathbb{R}^{512 \times 8 \times 8}$. The extracted features from both sides are then taken by a decoder $G_{1}$ to produce a reconstructed RGB-image $I^{ae}_g$. Note that the whole training process for our AE is on paired data after we synthesize multiple sketches for each image in the RGB-dataset using TOM.

\smallskip
\noindent\textbf{Translation-Invariant Style Encoder} To let $E_{style}$ extracts translation-invariant style information, thus approach a content-invariant property, we augment the input images by four image translation methods: cropping, horizontal-flipping, rotating, and scaling. During training, the four translations are randomly configured and combined, then applied on the original image $I_{rgb}$ to get $I^t_{rgb}$. Samples of $I^t_{rgb}$ drawn from an $I_{rgb}$ are shown on the top-left portion of Figure~\ref{fig:model-overview}, which $E_{style}$ takes one as input each time. We consider $I^t_{rgb}$ now possesses a different content with its style not changed, so we have an reconstruction loss between the decoded image $I^{ae}_{g}$ and the original $I_{rgb}$.

To strengthen the content-invariant property on $f_{style}$, a triplet loss is also leveraged to encourage the cosine similarity on $f_{style}$ to be high between the translations of the same image, and low between different images:
\begin{align}
    \mathcal{L}^s_{tri} = \text{max}(cos( f^t_{s} , f^{org}_{s} ) - cos( f^t_{s} , f^{neg}_{s} ) + \alpha, 0),
\end{align}
where $\alpha$ is the margin, $f^{t}_{s}$ and $f^{org}_{s}$ are feature vectors from the same image, and $f^{neg}_{s}$ is from a different random image. The translations on $I_{rgb}$ enforces $E_{style}$ to extract style features from an content-invariant perspective. It guides our AE learn to map the styles by the semantic meanings of each region, rather than the absolute pixel locations in the image.

\smallskip
\noindent\textbf{Momentum mutual-information minimization} A vanilla AE usually produces overly smooth images, making it hard for the style encoder to extract style features such as unique colors and fine-grained textures. Moreover, the decoder may rely on the content encoder to recover the styles by memorizing those unique content-to-style relations. 

Inspired by momentum contrastive loss \cite{he2020momentum}, we propose a momentum mutual-information minimization objective to make sure $E_{style}$ gets the most style information, and decouples the style-content relation on $E_{content}$. Specifically, a group of augmented images translated from the same image are treated as one unique class, and $E_{style}$ associated with an auxiliary classifier is trained to classify them. To distinguish different images, $E_{style}$ is enforced to capture as much unique style cues from each image as possible. Formally, $E_{style}$ is trained using cross-entropy loss:
 \begin{align}
    \mathcal{L}^s_{cls} = - \log ( \frac{ \exp( E^{cls}_{style}(f_{style})[label] ) }{ \sum_j \exp(E^{cls}_{style}(f_{style})[j])}) ),
\end{align}
where $E^{cls}_{style}(\cdot)$, implemented as one linear layer,  yields the class prediction vector and $label$ is the assigned ground truth class for $I_{sty}$.
 
While $E_{style}$ is predicting the style classes, we can further decouple the correspondence between $f_{style}$ and $f_{content}$ by implicitly minimizing their mutual-information:
\begin{align}
\nonumber
    \text{MI}(f_{style}, f_{content}) = \text{H}(f_{style}) - \text{H}(f_{style}|f_{content})
\end{align}
where $H$ refers to entropy. Since $\text{H}(f_{style})$ can be considered as a constant, we only consider $\text{H}(f_{style}|f_{content})$ and encourage that style information can hardly be predicted based on $f_{content}$. In practice, we make the probability of each style class given $f_{content}$ equal to the same value. The objective is formulized as:
 \begin{align}
 \label{eq:mi}
    \mathcal{L}^c_{cls} = \norm{ \text{softmax}(E^{cls}_{style}( f_{content} )) - v }^2,
\end{align}
where $v$ is a vector with each entry having the same value $\frac{1}{k}$ ($k$ is the number of classes). Note that we use average-pooling to reshape $f_{content}$ to match $f_{style}$. Eq.\ref{eq:mi} forces $f_{content}$ to be classified into none of the style classes, thus helps removing the correlations between $f_{content}$ and $f_{style}$.

\smallskip
\noindent\textbf{``Generative" Content Encoder} Edge-map to image synthesis possesses a substantial pixel alignment property between the edges from the input and the desired generated image. Instead, realistic sketches exhibit more uncertainty and deformation, thus requires the model to hallucinate the appropriate contents from misaligned sketch-lines. We strengthen the content feature extraction power of $E_{content}$ with a self-supervision manner using data augmenting. 

Firstly, we already gain multiple synthesised sketches for each image from TOM (with varied line straightness, boldness and composition). Secondly, we further transform each sketch by masking out random small regions, to make the lines dis-continue. An example set of $I^t_{skt}$ can be find in Figure~\ref{fig:model-overview}. Finally, we employ a triplet loss to make sure all the sketches paired to the same $I_{rgb}$ have similar feature-maps:
\begin{align}
    \mathcal{L}^c_{tri} =  \text{max}( d( f^t_{c} , f^{pos}_{c} ) - d( f^t_{c} , f^{neg}_{c} ) + \beta , 0), 
\end{align}
where $d(,)$ is the mean-squared distance, $\beta$ is the margin, $f^{t}_{c}$ and $f^{pos}_{c}$ are features from the sketches that correspond to the same $I_{rgb}$, and $f^{neg}_{c}$ is from one randomly mismatched sketch. Such self-supervision process makes $E_{content}$ more robust to the changes on the sketches, and enables it to infer a more accurate and completed contents from sketches with distorted and discontinued lines.

\smallskip
\noindent\textbf{Feature-space Dual Mask Injection} DMI is proposed in Sketch2art \cite{liu2020sketch} for a better content faithfulness of the generation to the input sketches. It uses the sketch-lines to separate two areas (object contours and plain fields) from a feature-map and shifts the feature values via two learnable affine transformations. However, DMI assumes the sketches aligns well to the ground truth RGB-images, which is not practical and ideal. Instead of the raw sketches, we propose to use $f_{content}$ to perform a per-channel DMI, as $f_{content}$ contains more robust content information that is hallucinated by $E_{content}$.

\smallskip
\noindent\textbf{Simplified Adaptive Instance Normalization} AdaIN is an effective style transfer module \cite{huang2017arbitrary}:
\begin{align}
\label{eq:adain}
    f'_c = \underset{c \times h \times w}{ \overset{1}{\mathrm{IN}} ( f_{c} )} \overset{2}{\times} \underset{c\times1\times1}{\sigma(f_{s})} \overset{3}{+} \underset{c\times1\times1}{\mu(f_{s})},
\end{align}
where $\mathrm{IN}$ is instance normalization, $\mu$ and $\sigma$ are the instance-wise mean and std. In spite of AdaIN's success on style transfer, its instance normalization (operation-1 in Eq.\ref{eq:adain}) usually causes droplet effects to models that are trained on large corpus of images \cite{karras2020analyzing}. To resolve the problem, we only preserve the \textbf{channel-wise multiplication} part (operation-2 in Eq.\ref{eq:adain}) in AdaIN, and abandon the $\mathrm{IN}$ and addition (operation-1 and 3 in Eq.\ref{eq:adain}). Such simplification turns out working great in our model.

\smallskip
\noindent\textbf{All objectives} Figure~\ref{fig:model-overview} stage-1 shows the overview of our AE. Via the proposed self-supervision training strategies, our encoders extract the disentangled features $f_{content}$ and $f_{style}$, and the decoder $G_1$ takes $f_{content}$ via DMI and applies $f_{style}$ via channel-wise multiplication to synthesis a reconstructed image.
The summed objective for our AE is:
\begin{align}
\label{eq:ae-sum}
    \mathcal{L}_{ae}= \quad &\mathbb{E}[ \norm{ G_1( E_s(I_{rgb}) , E_c(I_{skt}) ) - I_{rgb} }^2 ] \nonumber \\
                      & + \mathcal{L}^c_{tri} + \mathcal{L}^s_{tri} + \mathcal{L}^s_{cls} + \mathcal{L}^c_{cls},
\end{align}
where the first part in Eq.\ref{eq:ae-sum} computes the mean-square reconstruction loss between $I_{ae}$ and $I_{rgb}$. Please refer to the appendix for more discussions on why we choose AE over variational AE \cite{kingma2013auto}, and the implementation details on the revised DMI and simplified AdaIN.

\subsection{Revised Synthesis via Adversarial Training}
\begin{figure}
    \centering
    \includegraphics[width=0.8\linewidth,height=4.4cm]{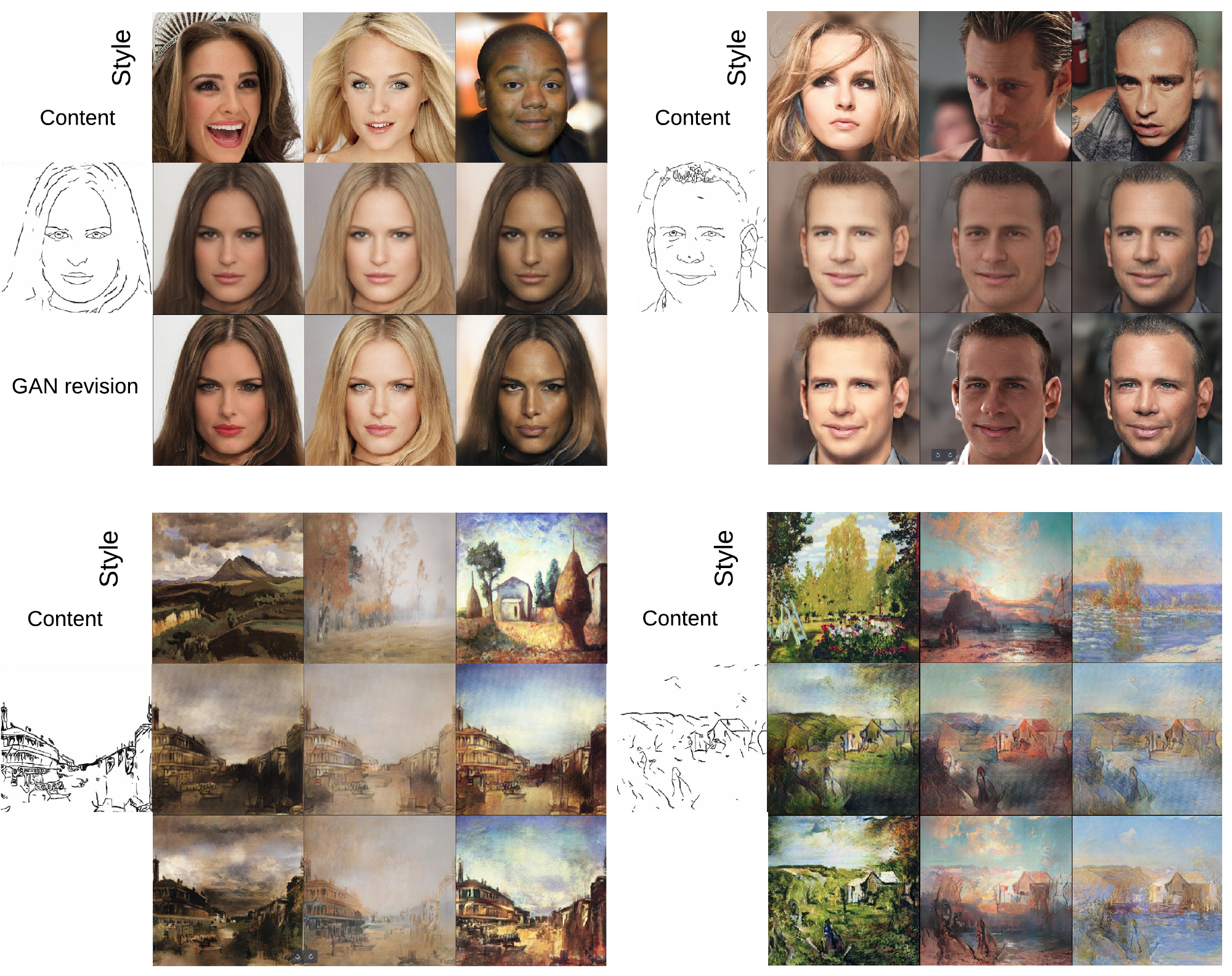}
    \caption{In each panel, the second row shows the images from AE, and the third row shows the GAN revisions.}
    \label{fig:ae2gan-images-1}
\end{figure}

Once our AE is trained, we fix it and train a GAN to revise AE's output for a better synthesis quality. As shown in Figure~\ref{fig:model-overview} stage-2, our Generator $G_2$ has a encoder-decoder structure, which takes $I^{ae}_{g}$ from $G_1$ as input and generates our final output $I^{gan}_g$. The final results of our model on unpaired testing data can be found in Figure~\ref{fig:ae2gan-images-1}, where $G_1$ already gets good style features and composites rich content, while $G_2$ revises the images to be much more refined. 

Same as our AE, only paired sketch and image data are used during the training. We do not randomly mismatch the sketches to images, nor do we apply any extra guidance on $D$. In sum, the objectives to train our GAN are:
\begin{align}
    \mathcal{L}_{D}= &-\mathbb{E}[ min(0, -1 + D(I_{sty})) ]  \nonumber \\
                     &-\mathbb{E}[ min(0, -1 - D( G_2( I^{ae}_g )) ], \\
    \mathcal{L}_{G_2}= &-\mathbb{E}[ D( G_2( I^{ae}_g)) ] \nonumber \\
                     &+ \lambda \mathbb{E}[ \norm{ G_2( I^{ae}_g ) - I_{sty} }^2 ],
\end{align}
which we employ the hinge version of the adversarial loss \cite{lim2017geometric,tran2017deep}, and $\lambda$ is the weight for the reconstruction term which we set to 10 for all datasets. Please refer to the appendix for more details.

\section{Experiments}
\textbf{Datasets} We evaluate our model on two datasets, CelebA-HQ \cite{liu2015faceattributes,CelebAMask-HQ} and WikiArt. 
\begin{itemize}
\item CelebA-HQ contains 30000 portrait images of celebrities worldwide, with a certain amount of visual style variance. We train our model on $1024^2$ resolution on randomly selected 15000 images and test on the rest images.
\item We collect 15000 high-quality art paintings from WikiArt \cite{wikiart}, which covers 27 major art styles from over 1000 artists. We train on 11000 of the images on $1024^2$ resolution and test on the rest images.
\end{itemize}

\subsection{Synthesis Sketches via TOM}
\begin{figure}[h]
    \centering
    \includegraphics[width=0.8\linewidth, height=5.8cm]{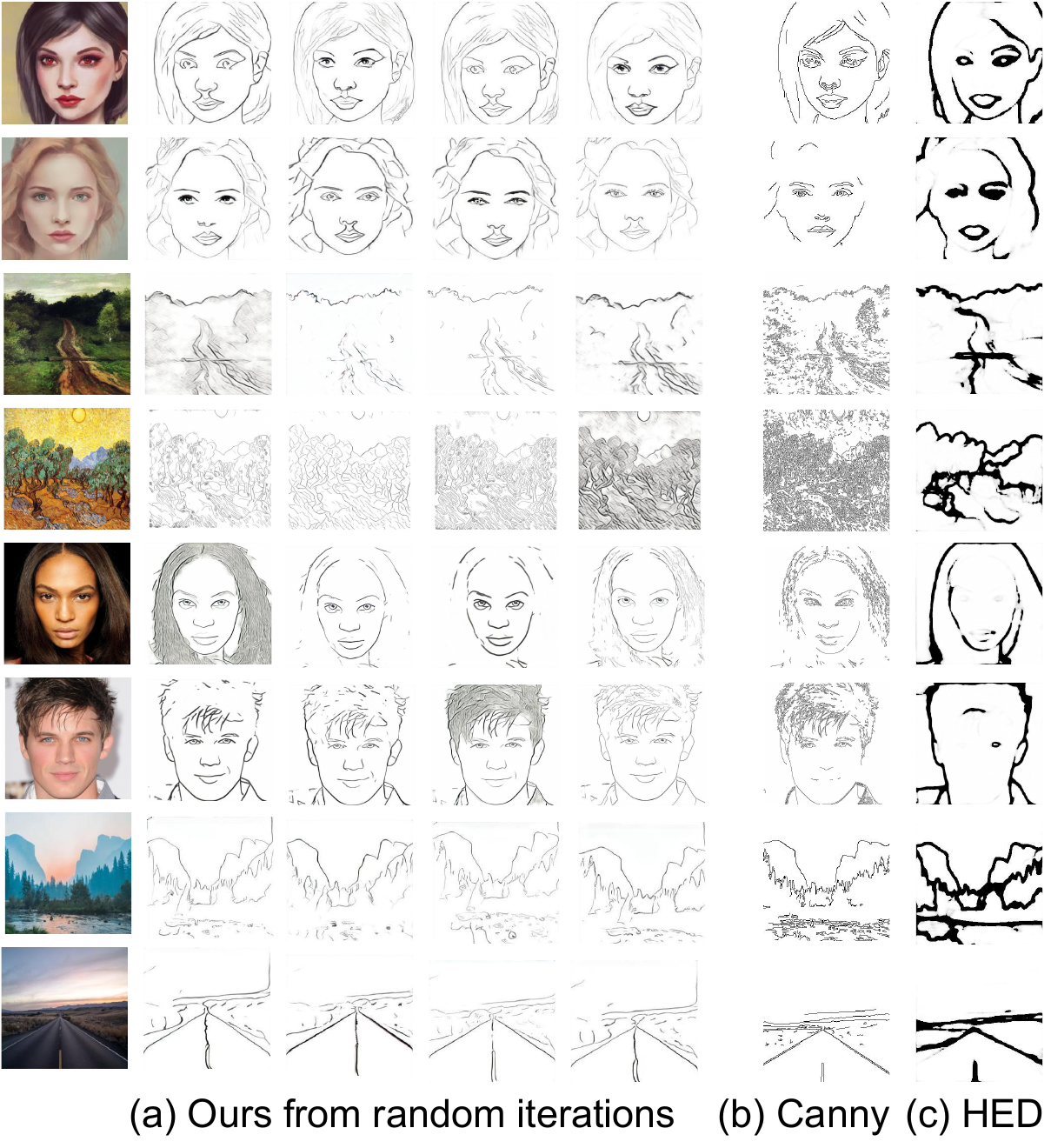}
    \caption{Synthesises from TOM. TOM generalizes well across multiple image domains, from photo-realistic to artistic, and from human portrait to nature landscape.}
    \label{fig:skt-sample}
\end{figure}

To train TOM, we find it sufficient to collect 20 sketches in the wild as $\mathcal{S}$. Moreover, the collected sketches work well for both the CelebA and WikiArt datasets. The whole training process takes only 20 minutes on one RTX-2080 GPU. We save ten checkpoints of $G_{sketch}$ to generate ten different sketches for an RGB-image. Figure~\ref{fig:skt-sample}-(a) shows the sketches generated from TOM. Among various checkpoints, we get sketches with diverse drawing styles, e.g., line boldness, line straightness, and stroke type. Moreover, while providing the desired sketch variations, it maintains a decent synthesis quality across all checkpoints. In comparison, edge detection methods are less consistent among the datasets. 

\subsection{Quantitative Evaluations}
\noindent\textbf{Quantitative metrics} We use three metrics: 
1) Fréchet Inception Distance (FID) \cite{heusel2017gans} is used to measure the overall semantic realism of the synthesized images. We randomly mismatch the sketches to the RGB-images and generate 40000 samples to compute the FID score to the real testing images.
2) Style relevance (SR) \cite{zhang2020cross} leverages the distance of low-level perceptual features to measure the consistency of color and texture. It checks the model's style consistence to the inputs and reflects the model's content/style disentangle performance. 
3) Learned perceptual similarity (LPIPS) \cite{zhang2018unreasonable} provides a perceptual distance between two images; we use it to report the reconstruction quality of our Auto-encoder on paired sketch and style image input.

\noindent\textbf{Comparison to baselines} We compare our model to the latest state-of-the-art methods mentioned in Section~\ref{sec:related-work}: RBNet (CVPR-2020), Sketch2art (SIGGRAPH-2020-RT-Live), CocosNet (CVPR-2020), and SPADE (CVPR-2019). Results from earlier methods, including Pix2pixHD, MUNIT, and SketchyGAN, are also presented.  Some models are adopted for exemplar-based synthesis to make a fair comparison and are trained on edge-maps as they originally proposed on. Instead, we train our model on synthesized sketches, which are more practical but arguably harder. We report the author's scores provided from the official figures, which, if not available, we try to train the models if the official code is published.   

\begin{table}[h]
\centering
\scalebox{0.85}{
\begin{tabular}{l|r r r r}
\toprule
 & \multicolumn{2}{c}{CelebA-HQ}  & \multicolumn{2}{c}{WikiArt} \\
 \cmidrule{2-5}
 & \multicolumn{1}{l}{FID $\downarrow$ } & \multicolumn{1}{l}{SR $\uparrow$ } & \multicolumn{1}{l}{FID $\downarrow$ } & \multicolumn{1}{l}{SR $\uparrow$ } \\
 \cmidrule{1-5}
Pix2pixHD  & 62.7  & 0.910  & 172.6 & 0.842 \\
MUNIT      & 56.8  & 0.911 & 202.8 & 0.856 \\
SPADE      & 31.5  & 0.941 & N/A   & N/A   \\
CocosNet   & 14.3  & 0.967 & N/A   & N/A   \\
SketchyGAN & N/A   & N/A   & 96.3 & 0.843 \\
RBNet      & 47.1 & N/A   & N/A   & N/A\\
Sketch2art & 28.7 & 0.958 & 84.2 & 0.897\\
 \cmidrule{1-5}
Ours AE       & 25.9  & 0.959 &  74.2 & 0.902 \\  
Ours AE+GAN   & \textbf{13.6}  & \textbf{0.972} & \textbf{32.6} & \textbf{0.924} \\
\bottomrule
\end{tabular}
}
\caption{Quantitative comparison to existing methods, bold indicates the best score. }
\label{table:baselines}
\end{table}

As shown in Table~\ref{table:baselines}, our model outperforms all competitors, and by a large margin on WikiArt. The self-supervised AE does a great job in translating the style features, while the GAN further boosts the overall synthesis quality.  

\begin{table}[h]
\centering
\scalebox{0.85}{
\begin{tabular}{c|c c}
\toprule
CelebA & FID $\downarrow$  & LPIPS $\downarrow$  \\
 \cmidrule{1-3}
Vanilla AE                   & 44.3  & 18.7  \\
AE + $\mathcal{L}^{c}_{tri}$ & 34.8  & 15.8  \\
AE + $\mathcal{L}^{s}_{tri}$ & 35.7  & 16.3 \\
AE + $\mathcal{L}^{s}_{cls}$ & 36.4  & 16.4 \\
AE + $\mathcal{L}^{s}_{cls}$ + $\mathcal{L}^{c}_{cls}$ & 34.7  & 15.2 \\
\cmidrule{1-3}
AE + all                     & 25.9 & 11.7 \\
\bottomrule
\end{tabular}
}
\caption{Benchmarks on the self-supervised objectives.}
\label{table:ae-objs}
\end{table}

\begin{figure}[h]
    \centering
    \includegraphics[width=0.9\linewidth]{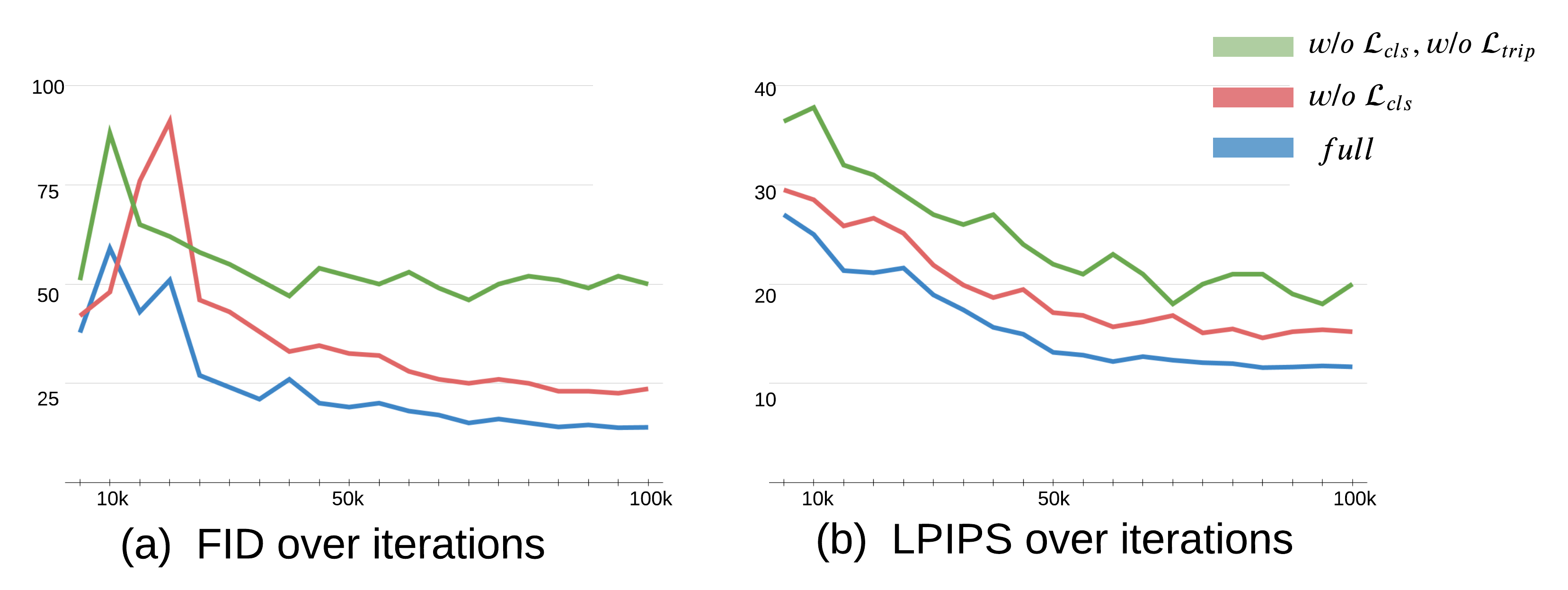}
    \caption{Model performance on CelebA during training.}
    \label{fig:ablation-scores}
\end{figure}

\noindent\textbf{Objectives Ablation} To evaluate the performance of AE, we compute FID from unpaired data to show its generalize ability, and compute LPIPS from paired data to show the reconstruction performance. Table~\ref{table:ae-objs} presents the contribution of each individual self-supervision objective. Compared to a vanilla AE with only reconstruction loss, each objective can independently boost the performance. Figure~\ref{fig:ablation-scores} better demonstrates the model behavior during training. We can see that the data-augmenting objectives $\mathcal{L}^{c/s}_{tri}$ make the biggest difference in the synthesis quality of the AE. Moreover, the contrastive objectives $\mathcal{L}^{c/s}_{cls}$ cooperates well with $\mathcal{L}^{c/s}_{tri}$ and further improves the scores.

\subsection{Qualitative Analysis}
A general sketch-to-image synthesis result of our model can be found in Figure~\ref{fig:samples-overview}. We select the style images that have a significant content difference to the sketches, to demonstrate the content/style disentangle ability of our model. Figure~\ref{fig:samples-overview}-(a) shows the result on WikiArt, which in a few examples, we still observe the ``content-interference from style image" issue, such as row.2-col.2 and row.7-col.3. Instead, on CelebA, as shown in Figure~\ref{fig:samples-overview}-(b), the model disentangles better even for rare style images such as col.4 and 5. This is expected as CelebA is a much simpler dataset in terms of content variance, whereas WikiArt contains much more diverse shapes and compositions.   

\begin{figure}
    \centering
    \includegraphics[width=0.9\linewidth]{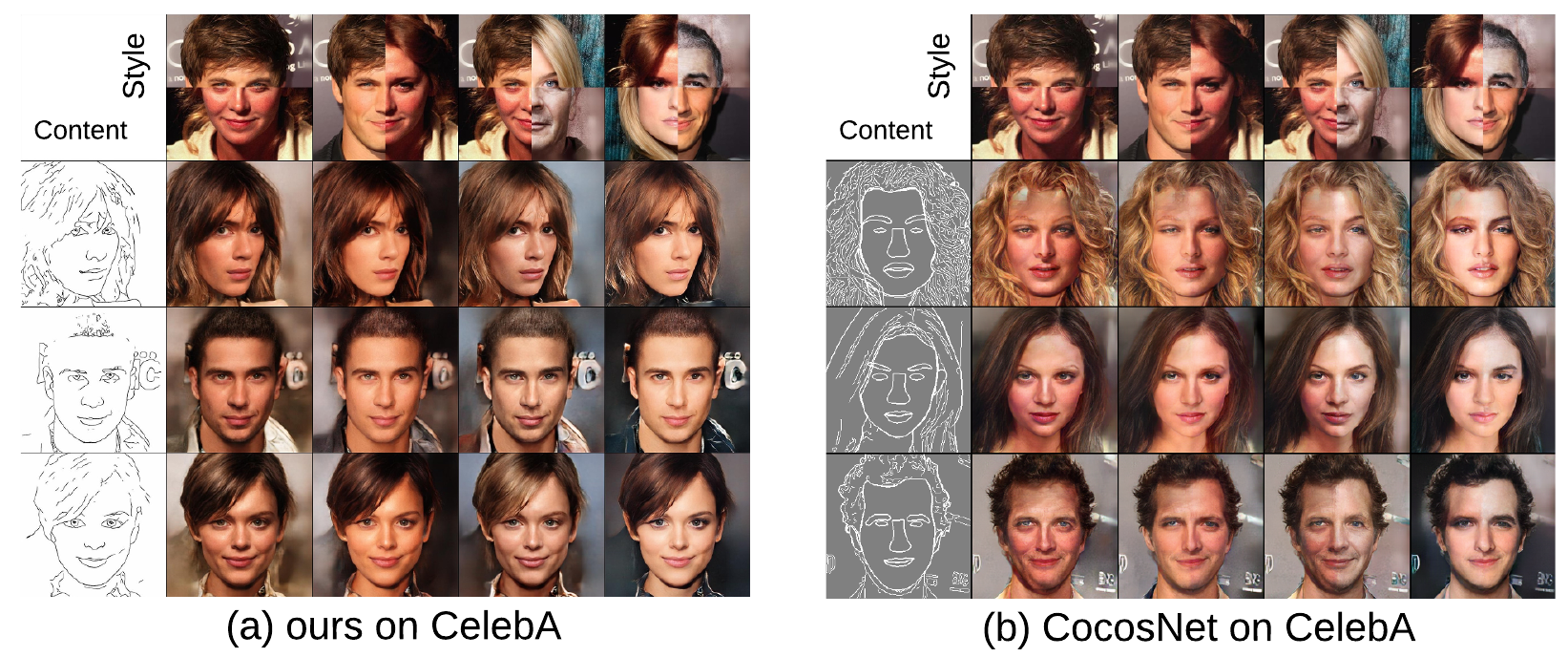}
    \caption{Synthesis by mixing multiple style images. }
    \label{fig:samples-mixing-styles}
\end{figure}

\noindent\textbf{Synthesis by mixing multiple style images} Via feeding structurally abnormal style images to the model, we demonstrate the model's superior ability on 1) capturing style cues from multiple style images at once; 2) imposing the captured styles to the sketch in a semantically meaningfully manner. Figure~\ref{fig:samples-mixing-styles} shows the synthesis comparison between our model and CocosNet on CelebA. We cut and stitch two or four images into one, and use the resulting image as the referential style. Our model harmonizes different face patches into unified style features, resulting in consistent hair color, skin tone, and textures. In contrast, CocosNet exhibits a patch-to-patch mapping between the input and output, yielding unrealistic color isolation on the synthesized images. Moreover, the color consistency of the style image on CocosNet is severely downgraded on mixed images, while our model summarizes a ``mixture style" from all patches. 

\begin{figure}[h]
    \centering
    \includegraphics[width=0.9\linewidth, height=9cm]{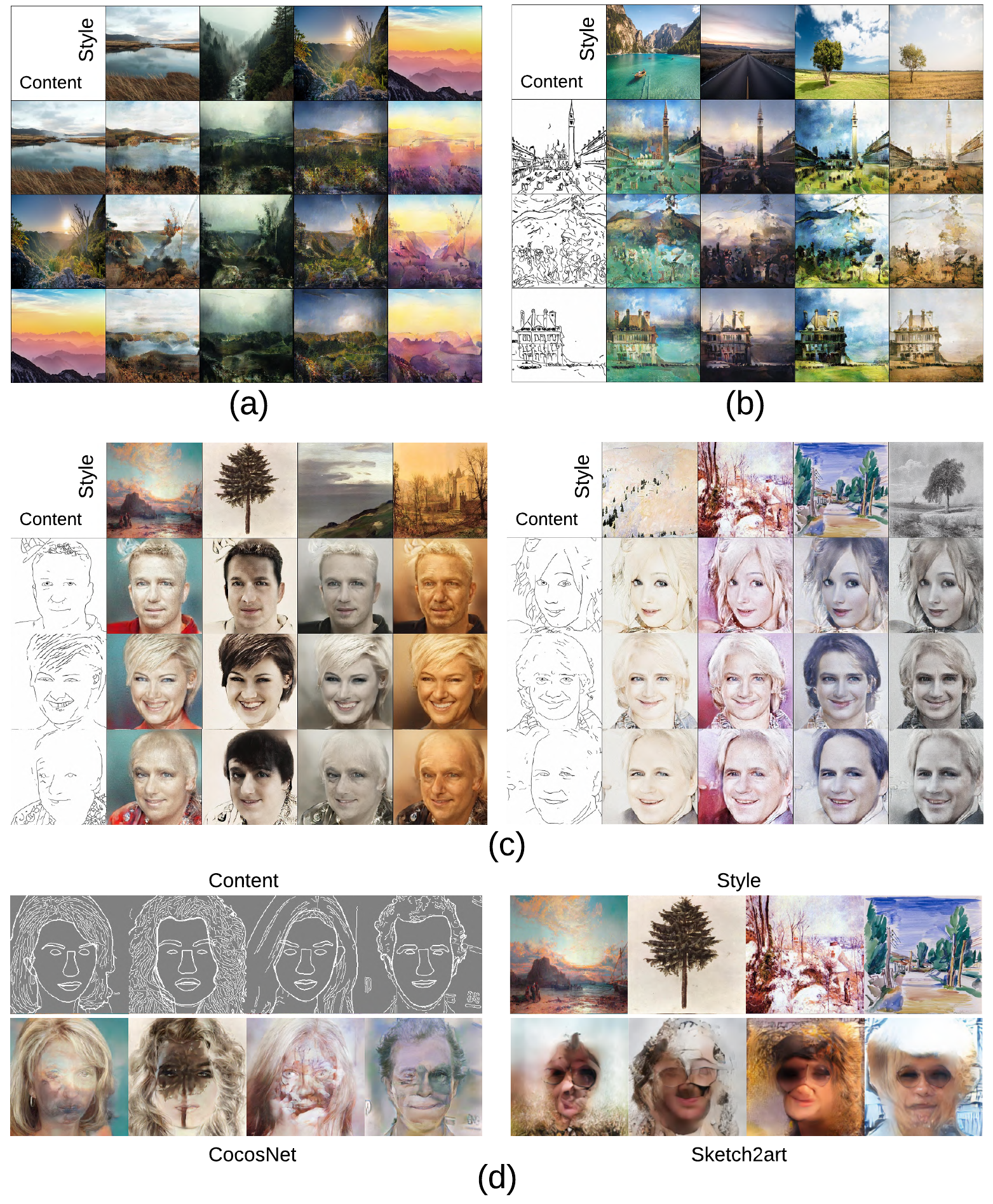}
    \caption{Synthesis from out-domain style images.}
    \label{fig:samples-out-domain}
\end{figure}

\noindent\textbf{Synthesis on out-domain images} To demonstrate the generalization ability of our model, we use images from a different semantic domain than what the model is trained on as style images or sketches. In figure~\ref{fig:samples-out-domain}-(a) and (b), we use style images from photo-realistic nature scenes on our model trained on Wikiart. In figure~\ref{fig:samples-out-domain}-(a), the sketches are also from photo-realistic images, as shown in col.1, which we synthesize via TOM (note that TOM is also only trained on Wikiart). Although the out-domain images are different in texture, the model still gets accurate colors and compositions from the inputs. In figure~\ref{fig:samples-out-domain}-(b) row 2 and 4, the buildings are adequately colored, showing an excellent semantic inference ability of our model. Interestingly, an artistic texture is automatically applied to all the generated images, reflecting what the model has learned from the WikiArt corpus. 

In figure~\ref{fig:samples-out-domain}-(c), we use the art paintings as style images for the model trained on CelebA. All the faces are correctly generated and not interfered with the contents from the style images, showing our model's excellent job in hallucinating the contents from the input sketches. Amazingly, the model knows to apply the colors to the content follow proper semantics. Note how the hair, clothes, and backgrounds are separately and consistently colored. In contrast, figure~\ref{fig:samples-out-domain}-(d) shows how the other models suffer from generalizing on out-domain style images. CocosNet exhibits sever content-interference issue from the style images, and Sketch2art can hardly synthesis a meaningful face.

\begin{figure}
    \centering
    \includegraphics[width=0.85\linewidth, height=8cm]{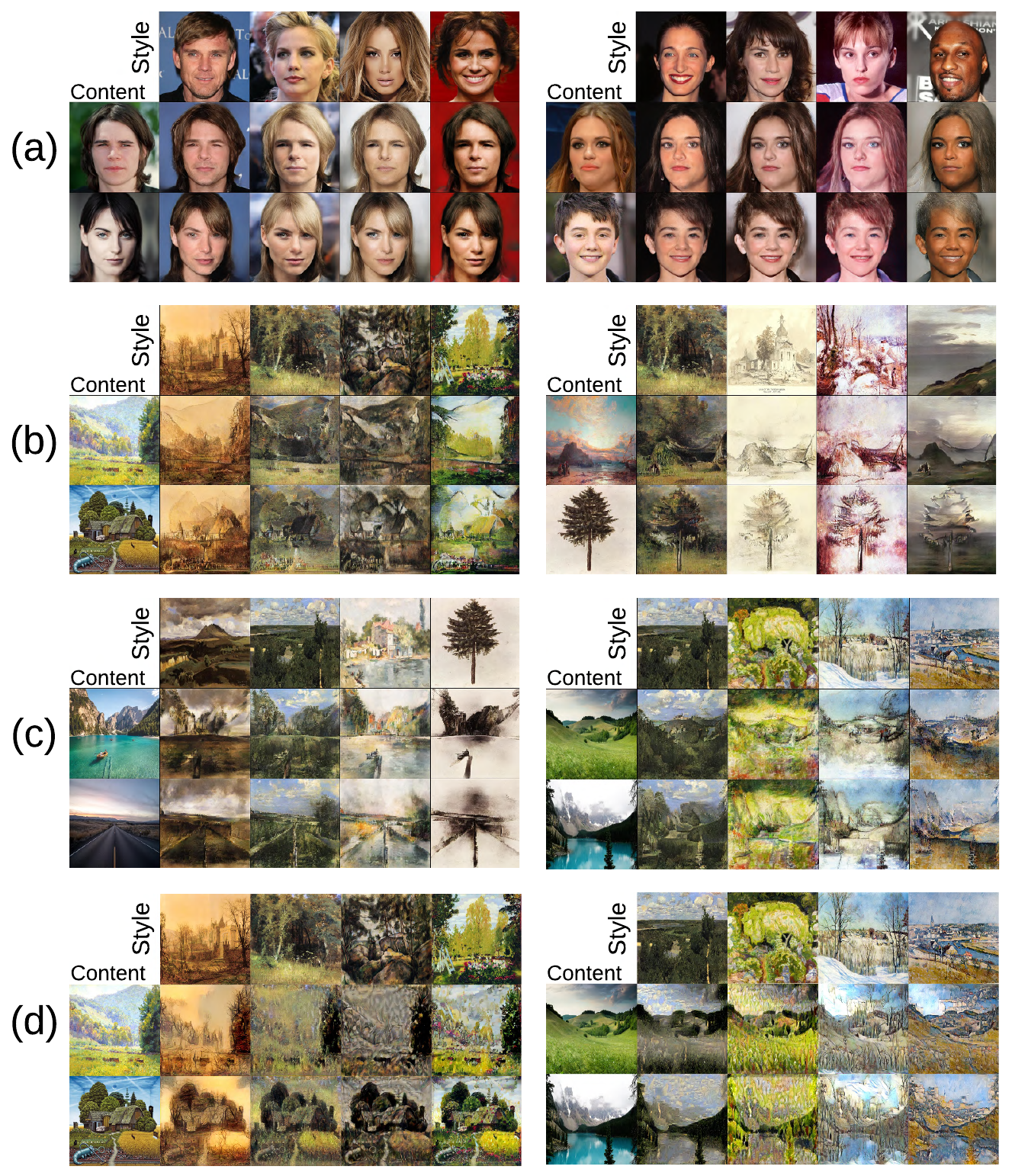}
    \caption{Works as a feed-forward style-transfer model.}
    \label{fig:samples-style-transfer}
\end{figure}

\noindent\textbf{Work as a style-transfer model} Combined with TOM, our model possesses competitive style-transfer ability. We first convert a content image into sketch-lines, then colorize it according to the style image. In this progress, our model can apply the style cues to different objects in the content image in a semantically appropriate manner. As shown in figure~\ref{fig:samples-style-transfer}-(a) and (b), our model can easily transfer the styles between in-domain images on face and art. Figure~\ref{fig:samples-style-transfer}-(c) further shows how the model performs on out-domain images. In contrast, Figure~\ref{fig:samples-style-transfer}-(d) shows the result of the traditional Neural Style Transfer (NST) method from \citeauthor{gatys2016image}, which an undesired texture covers the whole image in most cases. Note that we do not intend to compete with NST methods. Instead, our model provides a new perspective on the style transfer task towards a more semantic-aware direction.

Due to the space limitation, please refer to the appendix for more quantitative and qualitative experiments.

\section{Conclusion}
In this paper, we present a self-supervised model for the exemplar-based sketch to image synthesis. Without computationally-expensive modules and objectives, our model (trained on single GPU) shows outstanding performance on $1024^2$ resolution. With the mechanisms (self-supervisions) in this model orthogonal to existing image-to-image translation methods, even more performance boosts are foreseeable with proper tweaking and integration. Moreover, the extraordinary generalization performance on out-domain images showing a robust content and style inference ability of our model, which yields a promising performance on style-mixing and style-transferring, and reveals a new road for future studies on these intriguing applications.    

\noindent\textbf{Acknowledgement} At \url{https://create.playform.io/sketch-to-image}, demo of the model in this paper is available. The research was done while Bingchen Liu, Kunpeng Song and Ahmed Elgammal were at Artrendex Inc.

{\small
\fontsize{9pt}{10pt} \selectfont
\bibliographystyle{aaai21}
\bibliography{refs}
}
\clearpage

\appendix
\noindent\textbf{Appendix}
\section{Sketch-to-image model}

\subsection{Momentum contrastive loss}
Instead of predicting class labels on all images in the dataset at once, we conduct the ``momentum" training. During training, we randomly pick a small subset of $k$ images, and train the classification task only within this subset. For every constant amount of iterations, we randomly pick next subset of images and assign class labels, and re-initialize the weights in the model's final predicting layer. In practice, we find that setting k from 500 to 2000 yields similar performance and will not increase the computation burden by much. 

It is not desired to train the contrastive loss on all images. On one hand, the computation cost is increased; on the other hand, the performance boost effect is downgraded. This is because within a dataset, there are many images sharing a similar style. Force the model to predict different class labels on similar styles will lead to the model ignore the vital style information, and instead try to distinguish these images by remembering their different content information. 

\subsection{Feature-space Dual Mask Injection} DMI uses the lines of a sketch as an indicator (a feature mask) to separate out two areas (one around the object contours and one for the rest plain fields) from a feature-map, and shifts the feature values of the two areas with different learnable affine transformations. As a result, the shifted feature-maps lead to more faithful object shapes to the input sketch. We propose an improved version of DMI by using the feature-maps as the masks, rather than the single channel raw sketches. Interestingly, the improved DMI resembles a similar idea as spatially-adaptive normalization (SPADE) \cite{park2019semantic}, which relies on semantic labels for image synthesis. Instead, we are working on sketches without labeled semantic information, and our $E_{content}$ plays the role of inferring meaningful semantics.   

We conduct experiments on WikiArt to show the effectiveness of the proposed DMI, as WikiArt contains the images with the most complicated and varied compositions. On testing dataset, we compute LPIPS between the input style images (with paired sketches) and the reconstructed images, it shows how content-faithful the reconstructed images are following the sketches. We also compute a ``sketch reconstruction" (Skt-Rec) score on unpaired data, by matching the input sketch and the sketch extracted from the generated images using TOM. It provides more explicit indication on how well the generated image is following the content from the input sketch. The result can be found in Figure~\ref{table:dmi} which ``DMI" is the original module using raw sketches, and ``Feat-DMI" is the proposed feature-level DMI.  
\begin{table}[]
\centering
\begin{tabular}{@{}lll@{}}
\toprule
              & LPIPS & Skt-Rec \\ \midrule
AE            & 24.15 & 0.13    \\
AE + DMI      & 20.7  & 0.091   \\
AE + Feat-DMI & 18.6  & 0.075   \\ \bottomrule
\end{tabular}
\caption{DMI performance evaluation}
\label{table:dmi}
\end{table}

\subsubsection{Simplified Adaptive Instance Normalization} AdaIN \cite{huang2017arbitrary} is an effective style transfer module. It transfers the statistics from a target feature-map $f_s$ to a content feature-map $f_c$:
\begin{align}
\label{eq:adain}
    f'_c = \underset{c \times h \times w}{ \overset{1}{IN} ( f_{c} )} \overset{2}{\times} \underset{c\times1\times1}{\sigma(f_{s})} \overset{3}{+} \underset{c\times1\times1}{\mu(f_{s})}.
\end{align}
In spite of AdaIN's success on style transfer, its instance normalization (operation-1 in eq-\ref{eq:adain}) usually causes droplet effects to models that are trained on large corpus of images as discovered in \cite{karras2020analyzing,liu2020sketch}. To resolve the problem, we only preserves the \textbf{channel-wise multiplication} part (operation-2 in eq-\ref{eq:adain}) in AdaIN, and abandon the IN and the addition of the mean style vector to the feature-map (operation-1 and 3 in eq-\ref{eq:adain}). 

We argue that multiplication forces the model to learn meaningful style cues on all its own feature channels, while addition makes the model lazy and rely on what the added vector gives. For example, if we introduce the style information by addition only, the original feature-map $f_c$ can have all-zero values while still can inherent proper style information from $f_s$. Instead, if we involve the style information by multiplication, it requires all channels in $f_c$ must already have meaningful value, so $f_s$ can choose from the channels via the multiplication.  

In practice, we take the feature vector $f_{style}$ from our style encoder and multiply it to the high resolution ($64^2$ to $512^2$) feature-maps in decoder. Coincidentally, it resembles the excitation part in SENet \cite{hu2018squeeze}. While in SENet, the multiplication is viewed as an channel-wise attention that gives a weight to each channel in a feature-map, we show its effectiveness as a style selector.

Similarly, we do experiments on WikiArt to show the effectiveness of the simplified AdaIN as the diversified image styles among art paintings. On testing dataset, we compute LPIPS between the input style images (with paired sketches) and the reconstructed images to show the overall performance of the model. Then we compute a ``style reconstruction" (Sty-Rec) score on unpaired data using cosine similarity, by matching the input style image's style vector and the extracted style vector from the generated images using Style encoder. A more consistent style transferring performance should yield a closer style vector. The result can be found in Figure~\ref{table:adain}. We also compared the performance of using only channel-wise multiplication and using only addition. Note how multiplication outperforms addition in the testing; and while multiplication gets a similar style-reconstruction score, it outperforms AdaIN in LPIPS, which means it gives an overall higher image synthesis quality. 
\begin{table}[]
\centering
\begin{tabular}{@{}lll@{}}
\toprule
               & LPIPS   & Sty-Rec \\ \midrule
AE             & 24.15   & 0.871    \\
AE + AdaIN     & 22.3  & 0.921   \\
AE + Addition  & 23.1    & 0.897   \\
AE + Multiplication & 21.6  & 0.923   \\\bottomrule
\end{tabular}
\caption{Style transfer performance evaluation}
\label{table:adain}
\end{table}

\subsection{AE vs VAE}
We find that AE without variational inference \cite{kingma2013auto} already gives satisfied performance, thus we do not train our AE to match any priors. Importantly, VAE optimizes an \textbf{estimated} KL-divergence to an \textbf{assumed} prior, it strongly restricts the representation power of the model. We would rather grant the model the freedom than force it to fit into some wrong distributions. Especially, both the celebA and the WikiArt dataset are clearly not following normal distribution through analysis, e.g., in CelebA, the hair and skin color are biased towards white people, and in WikiArt, the art styles are biased on several famous artists as they have more works archived.

The difference between AE and VAE is that the ELBO objective from VAE can restrict the encoded feature space thus makes all the encoded feature vectors stay close to each other. The ultimate goal for VAE is to achieve a continuous feature space thus one can traverse the space and generate meaningful image at any point. However, it is well-known that the ELBO objective hurts the image reconstruction performance and limits the expressiveness of the encoded feature vectors. In our task, we do not need the continuity of the latent space, while the expressiveness of the style vectors is the most important. It gives us good reason to us AE rather than VAE as the main structure, and the superior performance of our model supports our decision.    

\subsection{GAN}
The performance of our model can be further boosted with a tweak on the Generator $G_2$ during the GAN training. Apart from $I^{ae}_{g}$, $G_2$ can also take as input the style vectors $f_{style}$ from $E_{style}$, to recover the detailed style cues that may missed in $I^{ae}_g$. In practice, we also add random noises $z$ to feature-maps at multiple resolutions in $G_2$, to make it more compatible at imitating the fine-grained texture distributions of the images. 

Several previous models \cite{isola2017image,zhu2017unpaired,kim2019u,chen2018sketchygan,Lee_2020_CVPR} (Pix2pix, UNIT) employ a joint training method, which they combine and AE and GAN, by treating the decoder as generator and using an extra discriminator to train the model. In sum, the decoder is trained by an reconstruction loss and an adversarial loss. Our proposed model can also work in such settings, which means we merge the two-stage training into one by using the discriminator in stage-2 to jointly train the decoder at stage-1. However, we find such joint training performs not as good as the two-stage training method. Firstly, joint training requires more computation capacity as an extra discriminator is used in the same time as AE. It means we have to use a smaller batch size given the limited computing resource. Secondly, it is hard to balance the adversarial loss and the reconstruction loss when training AE from scratch. Undesired hyper-parameter tuning are introduced if AE and GAN are trained together. Lastly, in our two-stage training, we can use another generator which takes the output from AE as input, and learn more details on what the AE could not learn. The two-stage training considerably improves the image synthesis quality, and compared to previous joint-training methods, are more stable to converge and robust to train.

\section{Qualitative results}
\begin{figure*}
    \centering
    \includegraphics[width=\linewidth]{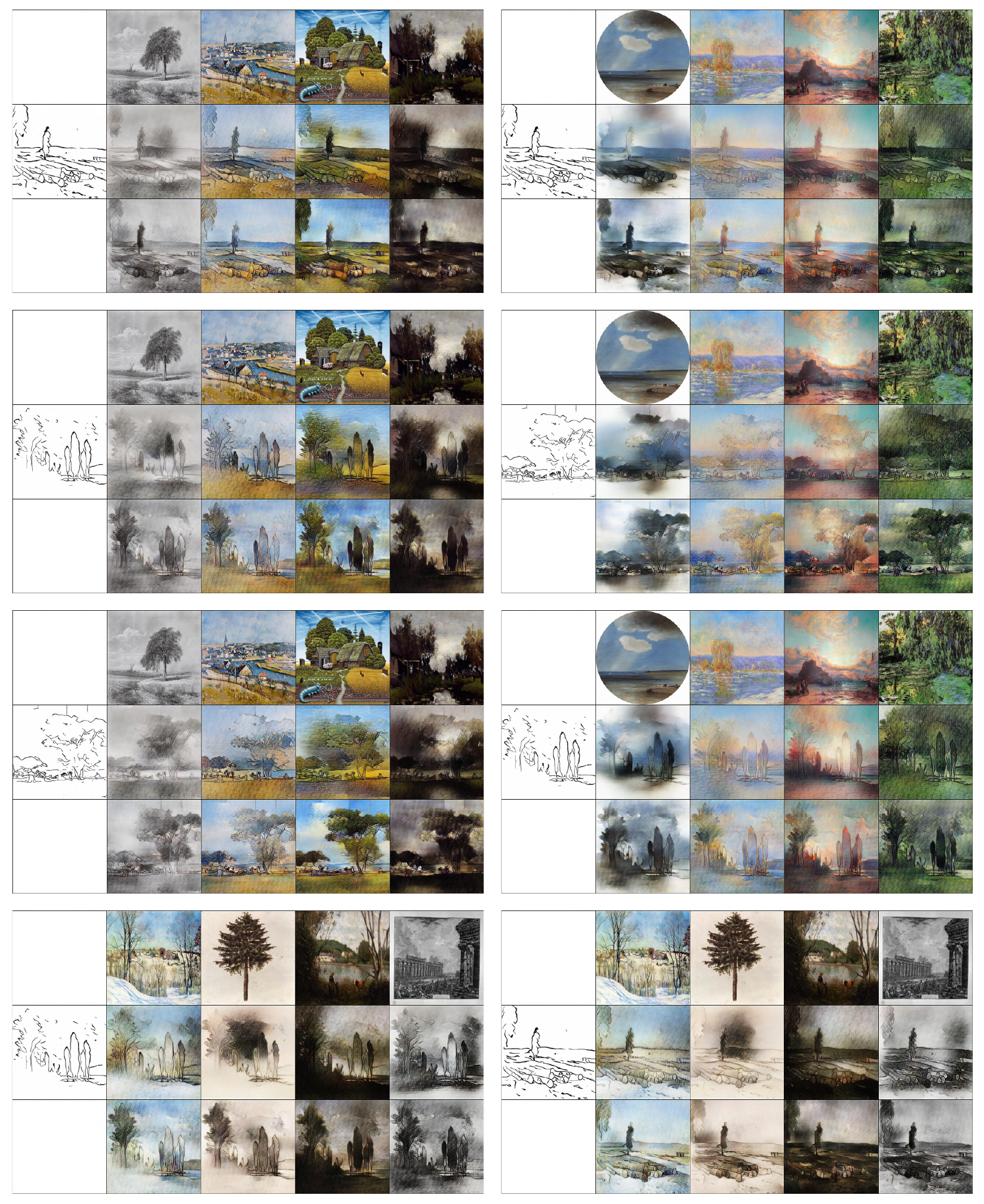}
    \caption{Qualitative comparison of the Auto-encoder without and with the proposed self-supervision objectives. In each panel, the first row are the referential style images, the first column is the input sketch, the second row are synthesis results from a plain AE trained with only reconstruction loss, the last row are the results from the proposed AE with self-supervision objectives: $\mathcal{L}^s_{tri}, \mathcal{L}^c_{tri}, \mathcal{L}^s_{cls}, \mathcal{L}^c_{cls}$. }
    \label{fig:ae_noss_compare_art}
\end{figure*}

\begin{figure*}
    \centering
    \includegraphics[width=\linewidth]{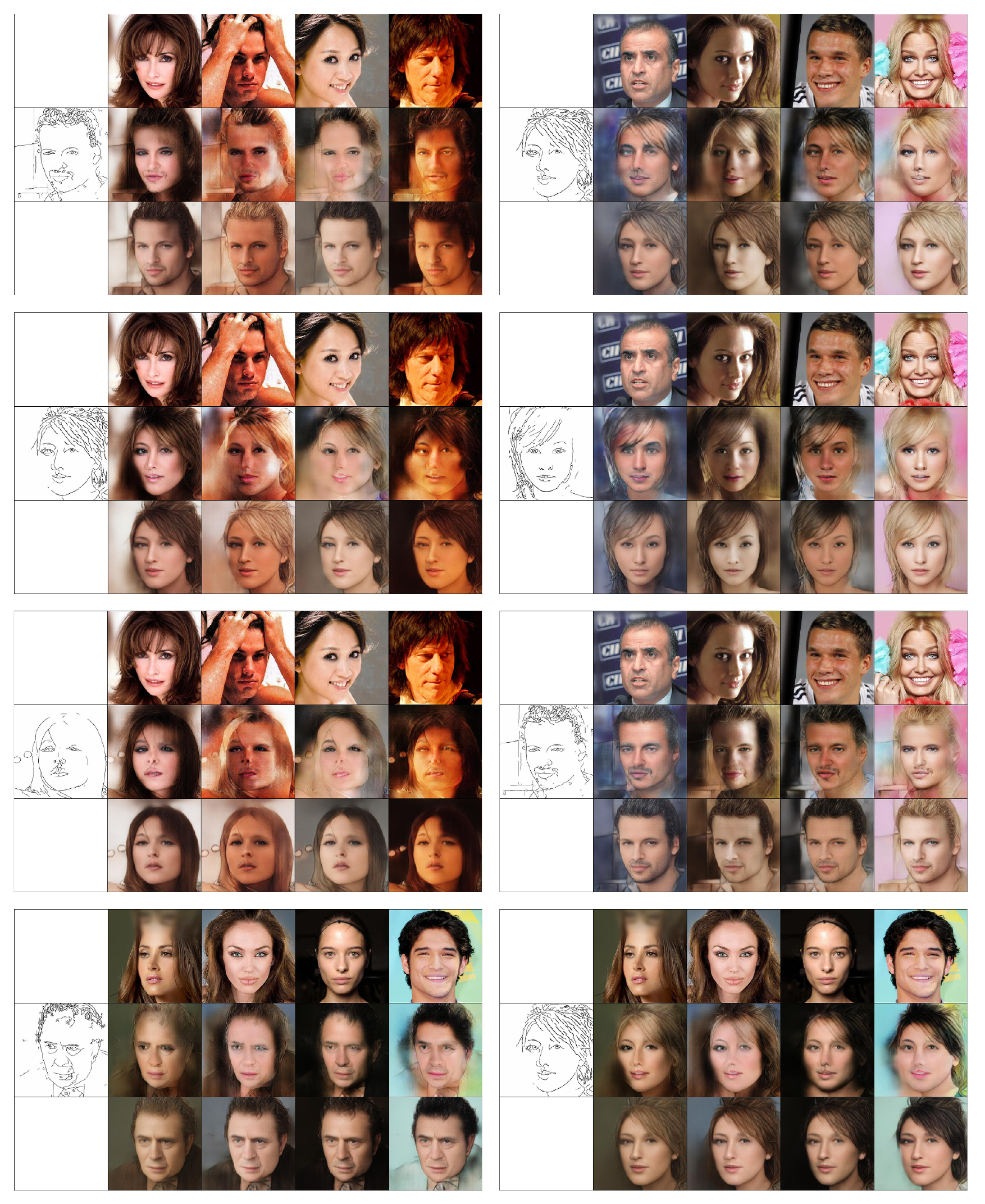}
    \caption{Qualitative comparison of the Auto-encoder without and with the proposed self-supervision objectives. The image arrangement is the same as Figure~\ref{fig:ae_noss_compare_art}}
    \label{fig:ae_noss_compare_face}
\end{figure*}

We present more qualitative results to provide a comprehensive idea of the performance of our model. As the main contribution of this paper is the self-supervised autoencoder, we compare the synthesis quality between the vanilla AE trained with only reconstruction loss and the proposed AE trained with self-supervised losses ( $\mathcal{L}^s_{tri}, \mathcal{L}^c_{tri}, \mathcal{L}^s_{cls}, \mathcal{L}^c_{cls}$ ) in Figure~\ref{fig:ae_noss_compare_art} and Figure~\ref{fig:ae_noss_compare_face}. The quality difference is rather obvious. Importantly, the ``content-interference from style image" issue is properly alleviated in our model. While for a vanilla AE, the generated images all exhibit a shadow imaginary of the style image, with undesired coloring on areas not indicated by the input sketch.

\begin{figure*}
    \centering
    \includegraphics[width=\linewidth,height=1.3\linewidth]{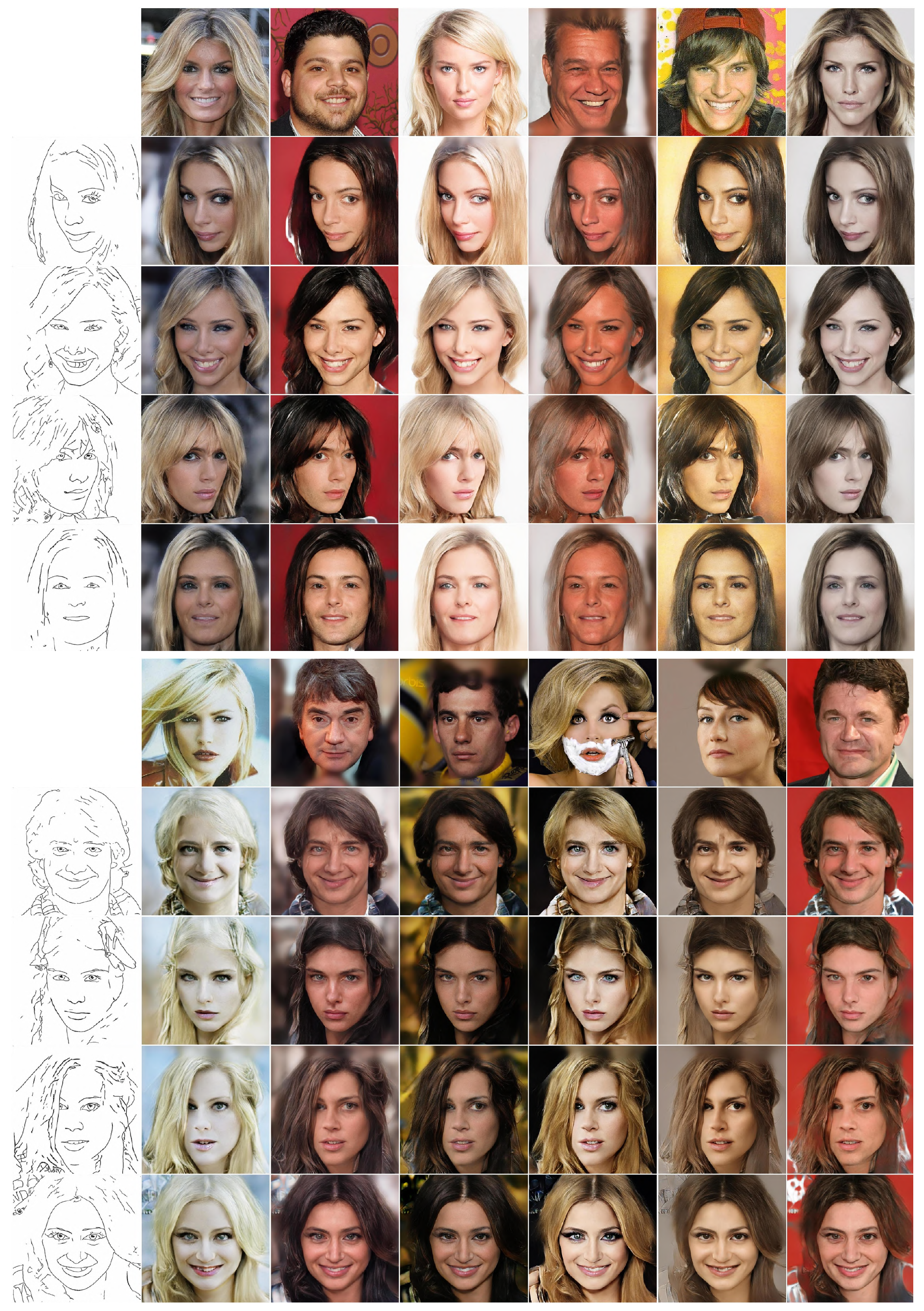}
    \caption{Uncurated synthesis results of our model on CelebA. Note that the images are compressed due to file size limit.}
    \label{fig:face_1}
\end{figure*}

\begin{figure*}
    \centering
    \includegraphics[width=\linewidth,height=1.3\linewidth]{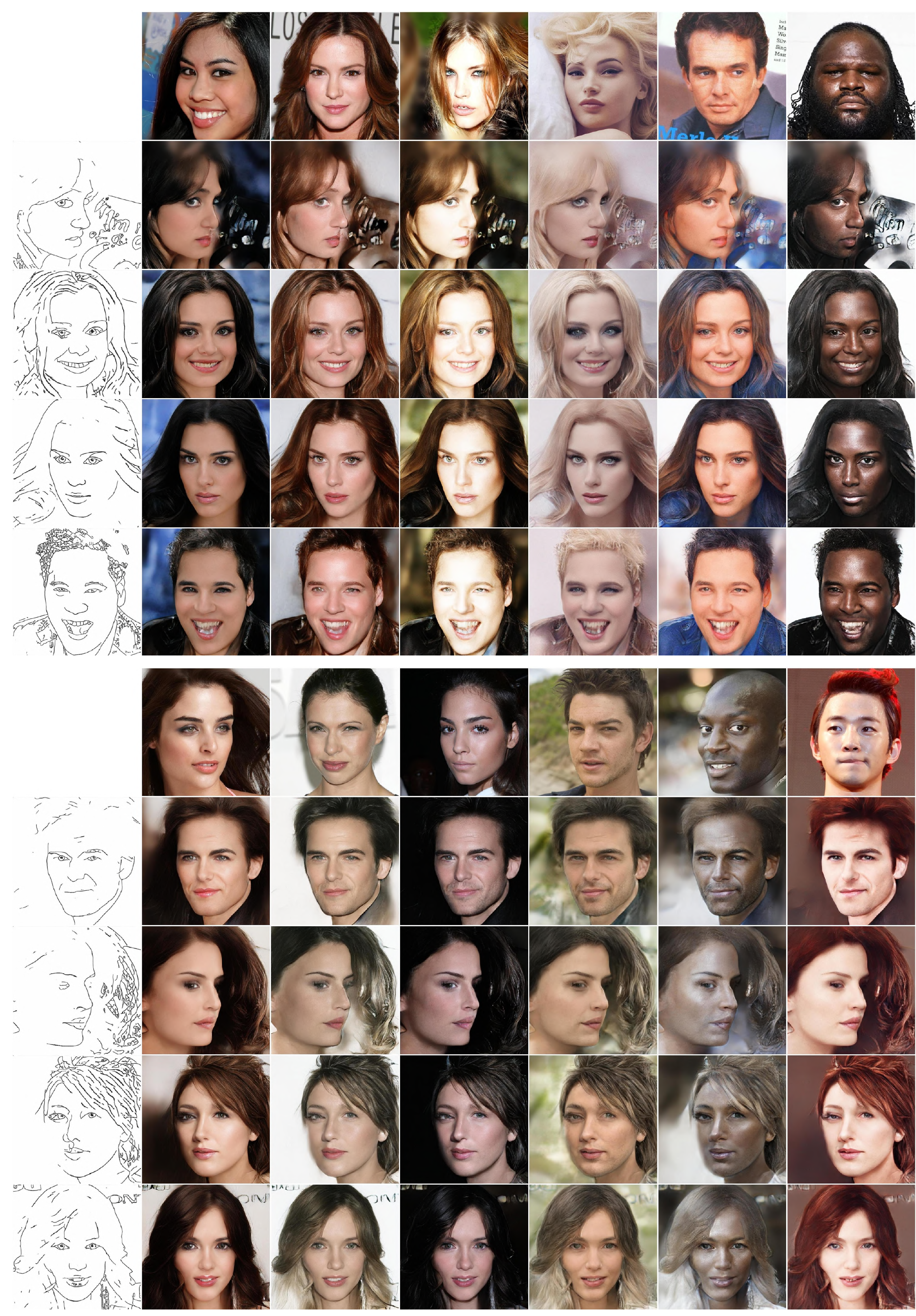}
    \caption{Uncurated synthesis results of our model on CelebA. Note that the images are compressed due to file size limit.}
    \label{fig:face_2}
\end{figure*}

\begin{figure*}
    \centering
    \includegraphics[width=\linewidth,height=1.3\linewidth]{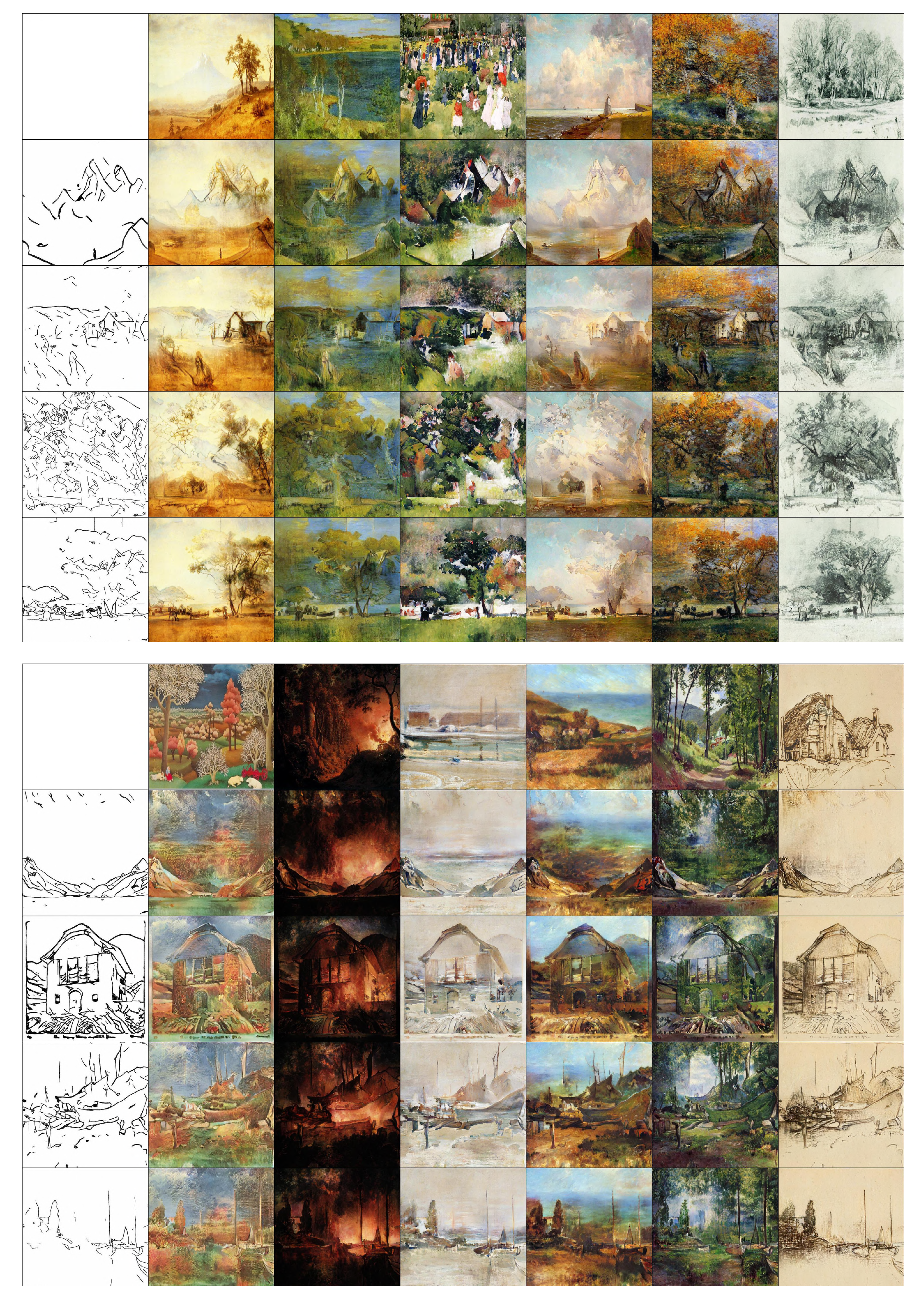}
    \caption{Uncurated synthesis results of our model on WikiArt. Note that the images are compressed due to file size limit.}
    \label{fig:art_1}
\end{figure*}

\begin{figure*}
    \centering
    \includegraphics[width=\linewidth,height=1.3\linewidth]{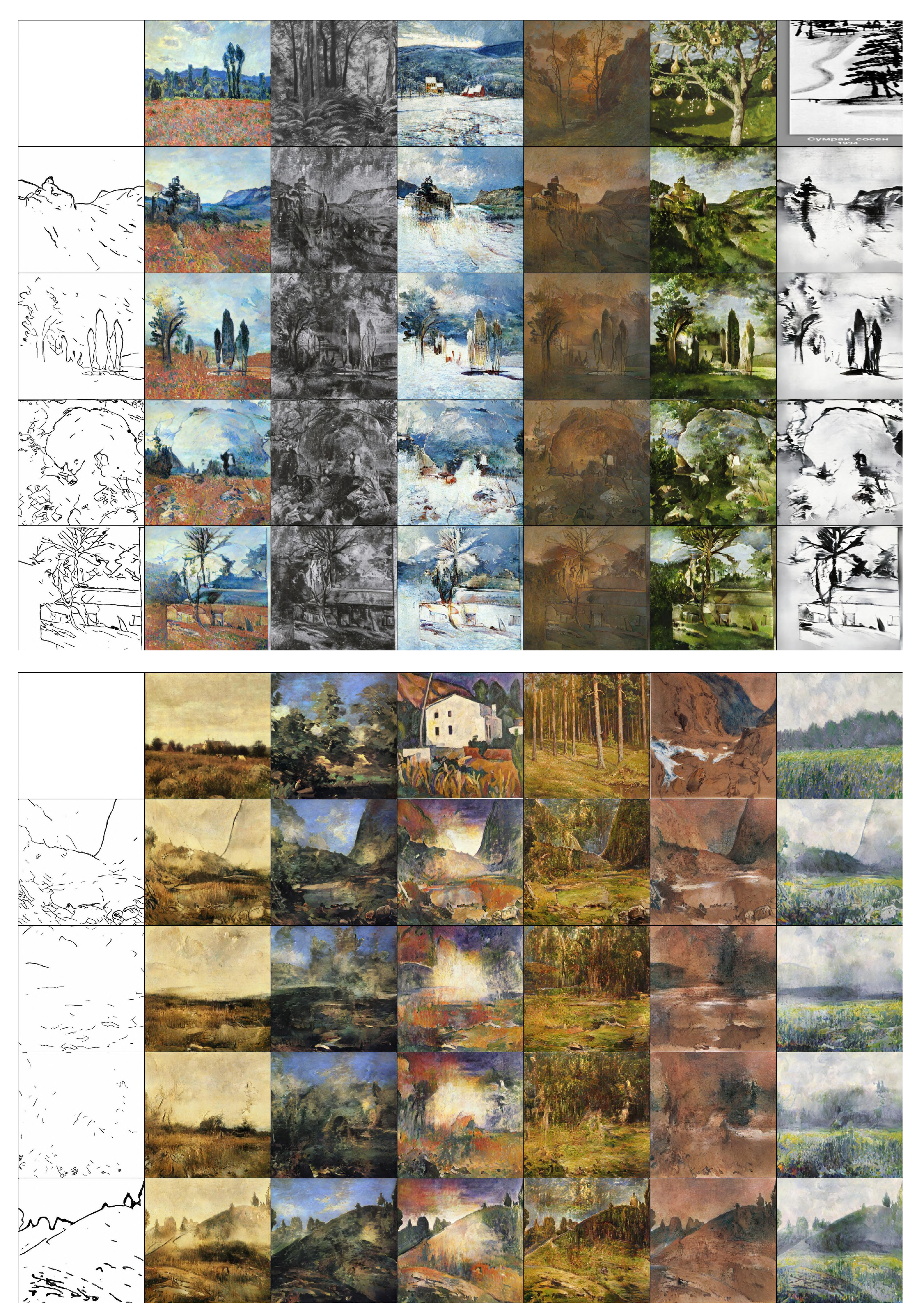}
    \caption{Uncurated synthesis results of our model on WikiArt. Note that the images are compressed due to file size limit.}
    \label{fig:art_2}
\end{figure*}

Figure~\ref{fig:face_1},~\ref{fig:face_2},~\ref{fig:art_1},~\ref{fig:art_2} shows more synthesized results from our model. We do not cherry pick the results, to better show the genuine performance of our model.  

\section{Implementation details}
We use the deep-learning framework PyTorch \cite{paszke2019pytorch} to implement our model, a completed code is provided in \url{https://github.com/odegeasslbc/Self-Supervised-Sketch-to-Image-Synthesis-PyTorch}, which is ready to run and reproduce our experiment results, the evaluation code is also provided, including computing FID and LPIPS. Please refer to the code for detailed model structures, training schemes, and data preparation procedures. 

The reported figures of our model in the paper is trained (both AE and GAN) on one Nvidia RTX Titan GPU, which has 24GB VRAM. We train AE for 100000 iterations with batch size of 16, and GAN for 100000 iterations with batch size of 12. The whole training requires 5 days, and we find that at 50000 (half) iterations for AE and GAN are able to generate appealing results (2.5 days). We find that training with two RTX Titan GPUs (which allows a larger batch-size) can further boost the performance, with the FID on CelebA boost to less than 10. Given the fact that previous methods (SPADE and CocosNet) training on 8 GPUs for the same days, our models outperforms them not only on performance but also on computation efficiency.

\subsection{Sketch target for TOM} 
\begin{figure}
    \centering
    \includegraphics[width=\linewidth]{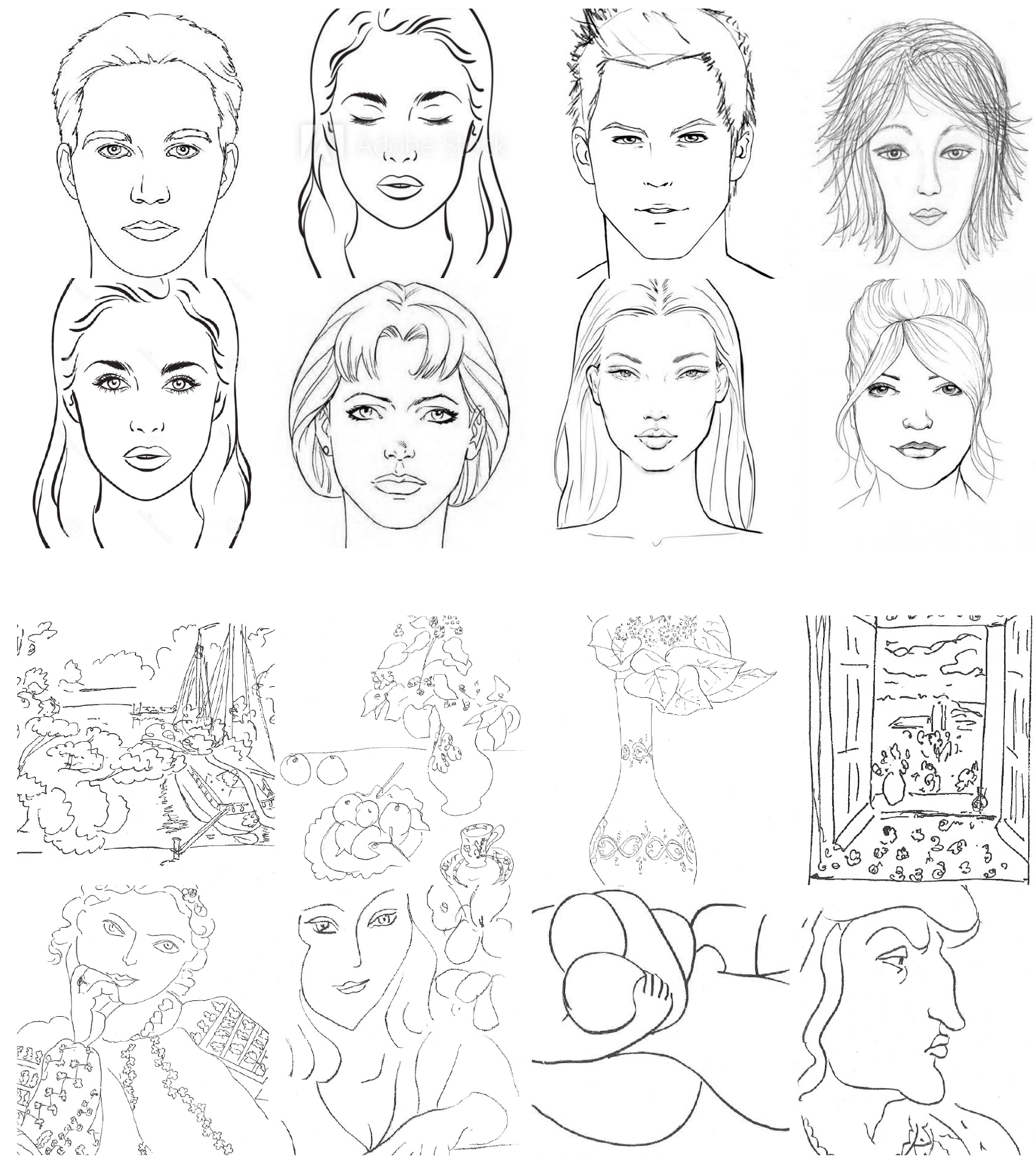}
    \caption{Example sketches we used to train TOM. The bottom panel are line-sketches from artists, and the top panel are sketches we randomly searched from the internet.}
    \label{fig:skt_target}
\end{figure}

Figure~\ref{fig:skt_target} shows the sketches we used as ground truth to train TOM for sketch synthesis on RGB-datasets. We collect in total of 60 images. Note that the same set of images are used for both the training on CelebA and WikiArt, indicating that the content domain of the ground truth sketches is not limited to be associated to the domain of the RGB-dataset.

\end{document}